%% file: smnist.tex
\documentclass[a4paper]{article}
\pdfoutput=1
\usepackage[utf8]{inputenc}
\usepackage[T1]{fontenc}
\usepackage{amsmath}
\usepackage{amssymb}
\usepackage{graphicx}
\usepackage[hyphens]{url}
\usepackage{cleveref}
\usepackage{graphicx}
\usepackage{caption}
\usepackage{subcaption}
\usepackage{csquotes}
\usepackage[export]{adjustbox}
\usepackage{authblk}
\usepackage{pgf,tikz,pgfplots}
\pgfplotsset{compat=1.14}
\usepackage{mathrsfs}
\usetikzlibrary{arrows}
\definecolor{feher}{rgb}{1,1,1}

\title{On the notion of number in humans and machines}

\author[1,*]{Norbert B\'atfai}
\author[2]{D\'avid Papp} 
\author[1]{Gerg\H{o} Bogacsovics}
\author[1]{M\'até Szab\'o}
\author[1]{Viktor Szil\'ard Simk\'o}
\author[1]{M\'ari\'o Bersenszki}
\author[1]{Gergely Szab\'o}
\author[1]{Lajos Kov\'acs}
\author[1]{Ferencz Kov\'acs}
\author[1]{Erik Szilveszter Varga}

\affil[1]{Department of Information Technology, University of Debrecen, Hungary}
\affil[2]{Department of Psychology, University of Debrecen, Hungary}
\affil[*]{Corresponding author: Norbert B\'atfai, batfai.norbert@inf.unideb.hu}

\begin{document}
\maketitle
\begin{abstract}
In this paper, we performed two types of software experiments to study the numerosity classification (subitizing) in humans and machines. Experiments focus on a particular kind of task is referred to as Semantic MNIST or simply SMNIST where the numerosity of objects placed in an image must be determined. The experiments called SMNIST for Humans are intended to measure the capacity of the Object File System in humans.
In this type of experiment the measurement result is in well agreement with the value known from the cognitive psychology literature. The experiments called SMNIST for Machines serve similar purposes but they investigate existing, well known (but originally developed for other purpose) and under development deep learning computer programs. These measurement results can be interpreted similar to the results from SMNIST for Humans. The main thesis of this paper can be formulated as follows: in machines the image classification artificial neural networks can learn to distinguish numerosities with better accuracy when these numerosities are smaller than the capacity of OFS in humans. Finally, we outline a conceptual framework to investigate the notion of number in humans and machines.
\end{abstract}


\section{Introduction}
\begin{figure}[h!]
    \centering
    \begin{subfigure}{.4\linewidth}
    	\centering
        \includegraphics[scale=2.5, frame]{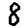}
        \caption{If the MNIST classifier takes this image of the digit 8 then it will say it's 8.}
        \label{sajat8a}
    \end{subfigure}
    \hskip2em
    \begin{subfigure}{.4\linewidth}
    	\centering
        \includegraphics[scale=2.5, frame]{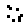}
        \caption{Provided the SMNIST classifier takes this image of 8 dots then it should say it's 8.}
        \label{sajat8p2}
    \end{subfigure}
    \caption{Two typical input images for MNIST and SMNIST.}
\end{figure}

In the movie Rain Man, there is a scene in which Dustin Hoffman as the autistic Raymond Babbitt can count the exact number of toothpicks on the floor in the blink of an eye. This scene gave the idea to implement it as a machine learning example. However to simplify this task we do not count toothpicks but dots in images. Let us compare this with the classical machine learning problem of recognizing MNIST  handwritten digit of numbers \cite{MNIST1}, \cite{MNISTHP}. In the classical MNIST task, a typical classifier program takes images of handwritten digits and recognizes them. An own image of the digit 8 can be seen in Fig. \ref{sajat8a}. 

The semantic MNIST, or shortly called SMNIST, program does not take images of digits but images that contain less than 10 dots. An image of 8 dots is shown in Fig \ref{sajat8p2}.

\subsection{Cognitive Neuropsychological and Computer Science Background and Aims}

The research of the biological and psychological factors behind the numerical abilities originate from the 1930-s. This extraordinary ability was studied from many angles. 
In insects, like the honeybee for example, there was found, that they can identify, and by that, count up to four different landmarks, for food reward \cite{Insect}.
In the neuropsychological literature there are two main topics of this type of research, which are the OFS (Object File System or its synonyms like Subitizing, Object Tracking System or Parallel Individuation System) and the ANS (Analogue Number System, Approximate Number System or also known as Analog Magnitude System) \cite{OFS}, \cite{Number}, \cite{PIS0}, \cite{PIS00}, \cite{PIS1}, \cite{ PISCoreSys2},   \cite{PIS2}.
The OFS includes the so-called \enquote{numerosity}, which is an ability, when someone just by looking at an object, without counting, can tell exactly how much of that given object is present. The maximum length of this ability in humans is up to four different objects, for a lifetime \cite{PIS0}, \cite{PIS00}. Therefore the OFS is a system, that helps us to determine the numerosity for a small quantity of items (maximum four), by using different markers for each object \cite{OFS}. There were many research made with vertebrates (like, for example, cats \cite{Cat}, \cite{Davis1988} chimpanzees \cite{Chimp}, \cite{Davis82} or parrots \cite{Parrot}), in which researchers studied the biological and evolutionary features for this particular ability. There are many research, that studied this ability in infants \cite{Child1}, \cite{Child2}, \cite{Child3}, \cite{Child4}, \cite{Child5}, \cite{Child6}) and proved the early, innate presence of numerosity \cite{PIS00}.
The ANS is a system that is present in a large scale of animals, and humans alike; this helps us to determine the numerosity of a small group of monitored objects, without using, or needing any kind of symbol- or language system. In the course of the biological development, this system is able to advance, and it can be looked at as a main foundation-stone for the progression of the numerical thinking \cite{Piazza1}, \cite{Piazza2}, \cite{Piazza3}.
In many cases mathematical simulation models have proven fruitful in cognitive neuropsychology research, for example \cite{NModel} planned a simulation of a natural neural architecture, where the distance \cite{dist} and size effect \cite{size} can be measured.
But pure mathematical models are not really rare either. For example \cite{GaussOFS} can explain the measured capacity of OFS.
The systems OFS and ANS of processing numbers clearly have evolutionary roots \cite{BrainNumbers}, \cite{BrainNumbers2}. In this light, it should be noticed that while they have presumably evolved over many hundreds of millions of years \cite{Number}, the mathematics has been developed over just a few thousand years.  Of course it is still possible that mathematics has evolutionary roots, see  the example about  Newton’s second law of \cite[1674]{Replicators} or Darwinian neurodynamics \cite{DND}.

From the viewpoint of computer science, the numerical abilities of computers are of analogue or digital nature \cite{Neumann}. In today's digital computers, numbers are represented in either fixed point or floating point format \cite{Knuth}.
Obviously, in contrast with previously cited neuropsychological systems, the numerical fundamentals of computers are fully known because they have been developed as results of targeted research and engineering processes as it has also been mentioned by McCulloch in \cite[319]{NeumannCW}. 
But it should be noticed that it will not be necessarily true for systems that include some deep learning black box AI \cite{BBAI} elements.
With this paper we would like to try to extend the above non-exhaustive listing of the cited works from vertebrates through human infants to include such items as that study numerosity classification in machine learning computer programs. In another context, this process has already begun. For example, see \url{https://rodrigob.github.io/are_we_there_yet/build/classification_datasets_results.html }
that presents the current state of the art in several standard machine learning tasks. The listed models and their implementations are typically based on artificial neural networks (ANNs) like convolutional neural network (ConvNet or CNN) \cite{LeCunConvNet}, \cite{ConvNet} or multilayer perceptron (MLP). We will run some of these well known, for example MNIST \cite{MNIST1} or CIFAR-10 \cite{cifar10}, programs in the second part of this paper. Nowadays, deep learning and artificial neural networks have already surpassed the human performance in several areas 
like, for example, playing old computer games \cite{ATARI}, 
playing GO \cite{GO}, playing Quake III Arena Capture the Flag \cite{Doom} or playing Starcraft \cite{AlphaStar0}. These three cited milestone works use reinforcement learning. There are many early roots of the success of these projects and deep machine learning in general. Such as the dataflow programming paradigm \cite{DataFlow}, the mathematical model of a neuron \cite{Neuron} or the concept of the perceptron \cite{Perceptron}. 
By now all key players in AI industry have their own frameworks for researching AGI (Artificial General Intelligence), for example, Microsoft uses MALMO-Minecraft \cite{MALMO}, Google uses DeepMind Lab-Quake III Arena  \cite{DeepMindLab} and so on, see \cite{Goliath}. The games serve as basis of these artificial environments are (or were) typically famous and popular computer games. 
Finally it should be noted that the Lamarckian evolutionary approach has already been arisen in this field as well \cite{LamEvol}, \cite{AlphaStar}.

The research experiments undertaken in this paper are divided into two main sections. The first one is the Semantic MNIST for Humans and the second one is the Semantic MNIST for Machines. While the aim of the experiment Semantic MNIST for Humans is clearly to investigate the capacity of OFS, the purpose of the experiment Semantic MNIST for Machines is less clear: we introduce a standard benchmark and several datasets of images for it.
We would like to investigate the numerosity classification abilities of deep learning programs that originally developed for other purpose.
A similar work can be found in \cite{DCNNNUM} where the notion of number in machines is investigated. 
If we compare it to our research we can see that our subitizing problem is simpler than theirs.     
Our more distant and utopian goal is to create a computer program that would be able to simulate the cognitive evolution of numbers, in the sense of Merlin Donald \cite{MerlinDonald}, and would be able to develop some kind of notion of number.

\section{Semantic MNIST for Humans}

The Android mobile application called \enquote{SMNIST for Humans} is a benchmark program intended to investigate the capacity of the parallel individuation system in humans. 
It is available in source code form from the GitLab project \cite{SMNISTREPO} under the directory \texttt{forHumans/SMNISTforHumansExp3}. But as a built APK file it can also  be downloaded and installed directly on Android devices from \url{http://smartcity.inf.unideb.hu/~norbi/SMNIST/SMNISTforHUMANS/Exp3/}.

\begin{figure}[h!]
    \centering
    \begin{subfigure}[t]{.24\linewidth}
\centering
\includegraphics[scale=.85]{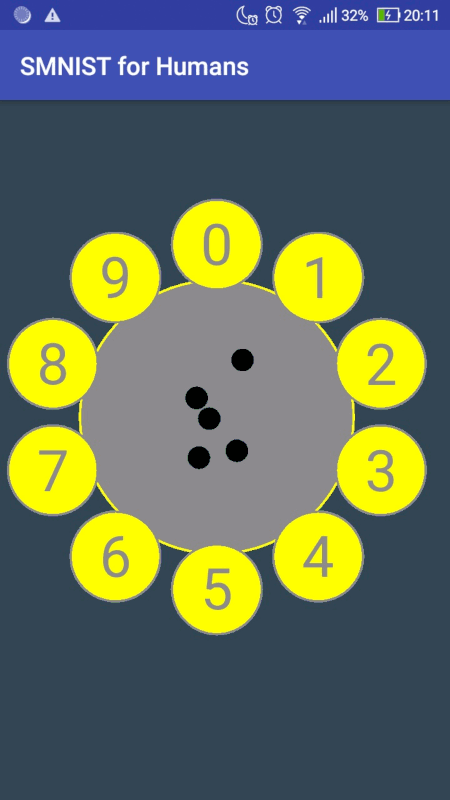}
\caption{A first rapid prototype for testing \enquote{gaming experience} with SMNIST for Humans. At each step of running the program, a random number of dots are drawn into the central circle which numerosity must be detected by touching the appropriate smaller circle of numerical digits.}
\label{smnistfh}
    \end{subfigure}
    \hskip2em
    \begin{subfigure}[t]{.56\linewidth}
\centering
\includegraphics[scale=.105]{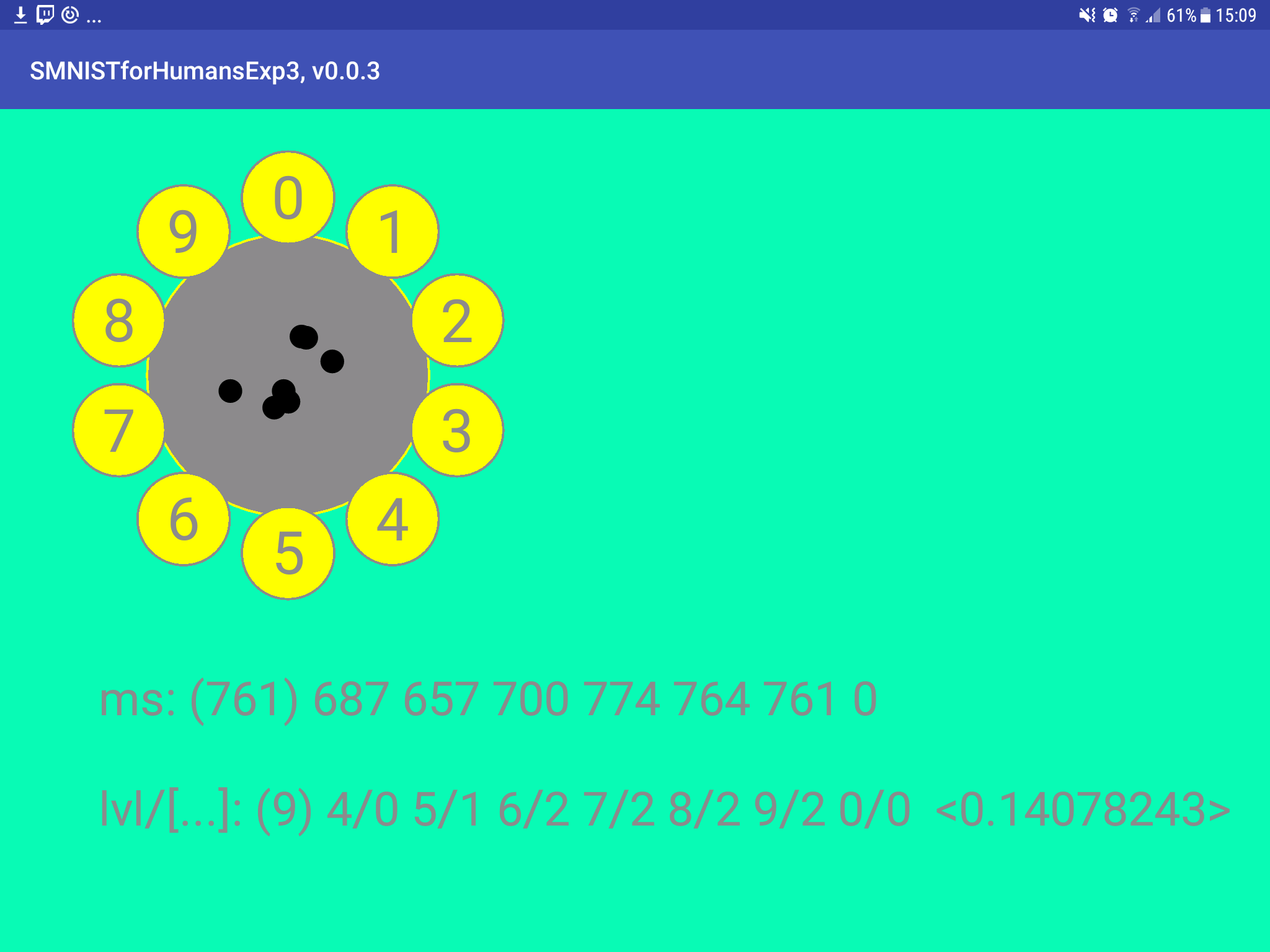}
\caption{This figure shows a screenshot of ``SMNIST for Humans, Experiment 3'' edition in action. The program displays (below in the second row) the changing of levels and the mean of numerosities of dots in addition (in the first row) the millisecond values corresponded to the changing of levels. Further precise details can be found in text.}
\label{bnlvl9}
    \end{subfigure}
    \caption{SMNIST for Humans screenshots.}
\end{figure}

As it can be seen in Fig. \ref{smnistfh}, the program draws a given number of dots on the screen then the user must touch the appropriate numerical digit within a certain time window. The players start at level 3 where 0, 1 or 2 dots can appear randomly on the screen. If players can detect the right number of dots for 10 consecutive times then they will move to the next level of the benchmark program. The achieved levels are indicated in the second row of numbers shown in Fig. \ref{bnlvl9}. Here the (9) 4/0 5/1 6/2 7/2 8/2 9/2 0/0 <0.14078243> row tells that the actual level (between round brackets) is 9. The 4/0 indicates that at the event of changing level from 3 to 4, 
the integer part of the mean of the 
randomly picked 10 (consecutive successfully detected) integers (numerosities of dots) was 0.
This is possible, for example, if the ten consecutive successfully detected numerosities are the following respectively 0, 0, 2, 1, 1, 1, 0, 1, 2, 1 where the integer part of the mean (0+0+2+1+1+1+0+1+2+1)/10 is equal to 0.
At changing from level 4 to 5 it was 1, from level 5 to 6 it was 2 and so on. 
Finally, the 0/0 shows that the player has not unlocked the level 10 yet. 
The last value between angle brackets is equal to the heuristic value  $\sum_{i=3}^{level}\frac{(l_i+1)(i+1)}{s_i}$ where $l_i$ denotes the mean of numerosities of dots of i-th level changing and $s_i$ is the millisecond value corresponded to the level changing. These millisecond values are displayed in the first row of numbers. The computed heuristic value serves only as a simple gamification element of the benchmark program. The greater this value the greater the performance of the player.

\begin{figure}[h!]
\centering
\includegraphics[scale=.5]{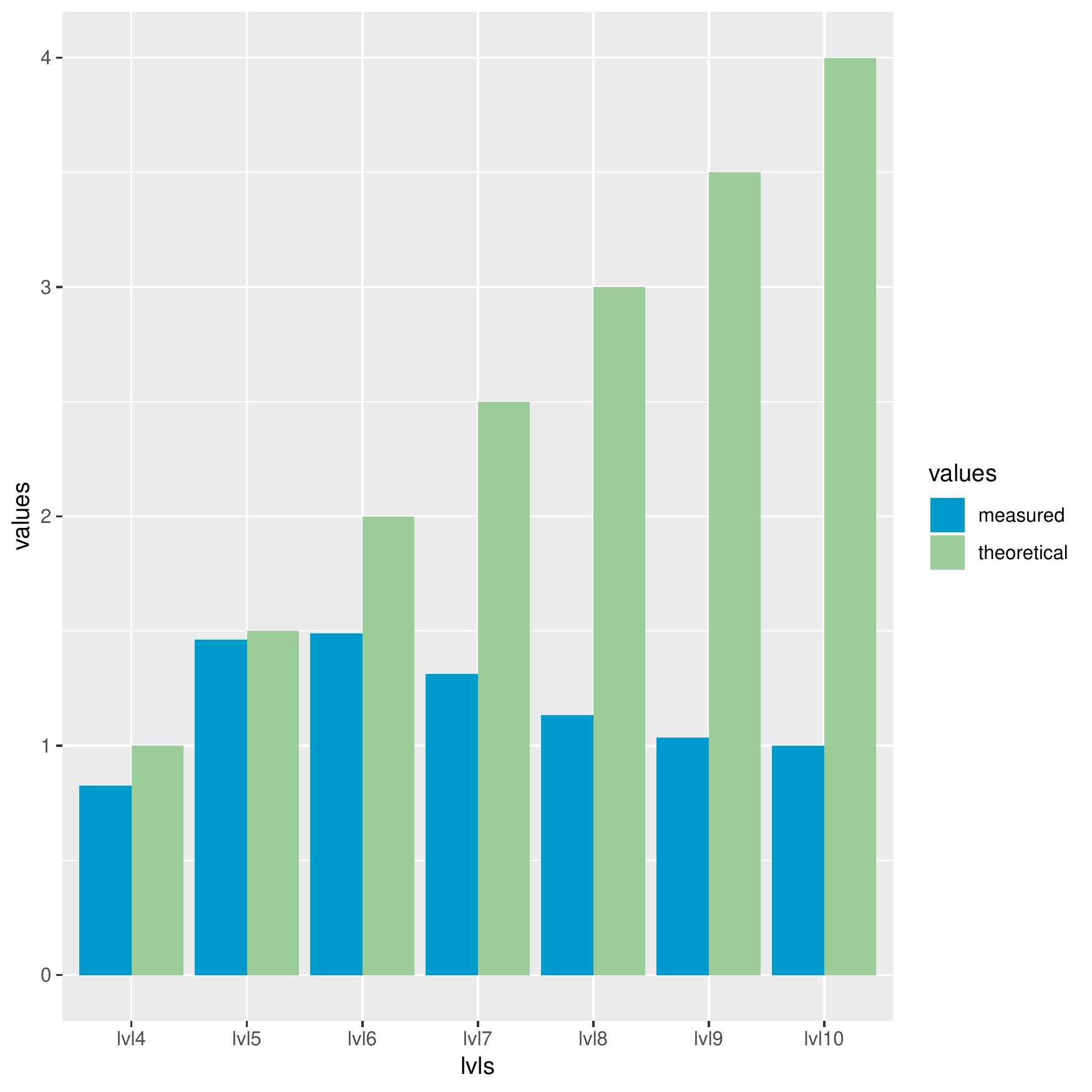}
\caption{This figure shows the relationship between the theoretical and measured mean of number of dots. The label lvl\{n\} denotes the event of changing level from n-1 to n. For example level 4 means that 0, 1 or 2 dots had already been successfully detected (and now the player is playing with 0, 1, 2 and 3 number of dots). The theoretical value denotes the expected value of the mean of randomly picked 10 integers from 0 to level-2, inclusive, that is $r_i \in [0, level-2]$ For example, it is equal to $(1+2)/3$ at level 4 or at level 10, it is equal to $(1+2+3+4+5+6+7+8)/9$, and in general it equals $(level-2)/2$. The measured value denotes the mean of the integer parts of means of the 10 consecutive successfully detected integers.}
\label{smnist-for-humans-exp3}
\end{figure}

Fig. \ref{smnist-for-humans-exp3} shows our measurement results. These results are in accordance with the well known observations from the cognitive psychology literature \cite{PIS0}, \cite{PIS1}, \cite{ PISCoreSys2},  \cite{PIS2} that the capacity of the parallel individuation system in humans is smaller than 4. It is demonstrated well by Fig. \ref{smnist-for-humans-exp3} where the measured average of the integer parts of means of randomly picked (and of course consecutive successfully detected) 10 integers is lagging far behind the theoretical expected value of the mean of randomly picked 10 integers from 0 to level-2, inclusive, that simply grows linearly with level, namely it is equal to $(level-2)/2$ (where $level$ starts from 4) because we use uniform distribution that is $r_i \in [0, level-2]$, $E(\sum_{i=1}^{10}r_i/10) =  \sum_{i=1}^{10}E(r_i)/10 = \sum_{i=1}^{10}(level-2)/20$. In this interpretation of levels the level n denotes the event of changing level from n-1 to n. For example level 4 means that 0, 1 or 2 dots had already been successfully detected (and now the 0, 1, 2 and 3 values are being  selected randomly).

Data were collected in the closed Facebook group of 680+ actual and former students of the BSc course “High Level Programming Languages” at the University of Debrecen called UDPROG where the students posted their results as  screenshots. A total of 104 Android device screenshots were received. One such screenshot can be seen in Fig. \ref{bnlvl9}.

\section{Semantic MNIST for Machines}

The SMINST for Machines is an attemption to develop a standardized task for assessing the ability of computer programs to recognize the numerosity of dots in an image. In the case of SMNIST for Humans, it is obvious that we do not need training dataset, but only test data (the random dots) that can be generated online during the running of the \enquote{game}. In contrast, in the case of SMNIST for Machines, we need both training and test datasets.

\subsubsection{The Generator Program}

The SMNIST datasets used in this study are generated by own generator program. It and its variants generate images that contain less than 10 dots. Their output are fully  binary compatible with the format of the original MNIST training and test data \cite{MNISTHP} so we can immediately start the first experiments using the former MNIST programs. 

\subsubsection{SMNIST datasets}

The datasets are organized into two releases (namely SMNIST for Machines and SMNIST for Anyone) and two series per release according to their development. 
The first series of SMNIST for Machines contains six pairs of train and test sets of $28x28$ images with the following properties
\begin{itemize}
\item Naive: this is a set of 60.000 training and 10.000 test $28x28$ pixels images that contain less than 10 randomly placed then centered dots. Dots are $3x3$ pixels.  
The histograms of the generated train and test images are 
0: 6025,
 1: 5977,
 2: 5965,
 3: 5928,
 4: 6075,
 5: 6067,
 6: 6004,
 7: 5930,
 8: 6051,
 9: 5978
and
  0: 986,
  1: 1008,
  2: 980,
  3: 963,
  4: 1064,
  5: 970,
  6: 996,
  7: 1010,
  8: 1036,
  9: 987.
\item No-centering: this set is generated by the same method as the previous one, but the randomly placed dots on the images are no centered. 
\item Disjunct: in this set all generated random images are unique images. It follows that training images are excluded from the test images. (Except the special case of 0 dots because there is just only one such image. It occurs several times in both sets.)
The histograms of the generated train and test images are 
0: 436,
1: 436,
2: 7390,
3: 7166,
4: 7482,
5: 7491,
6: 7299,
7: 7352,
8: 7498,
9: 7450
and  
0: 49,
1: 48,
2: 1215,
3: 1213,
4: 1227,
5: 1263,
6: 1282,
7: 1210,
8: 1209,
9: 1284.
\item Disjunct 1PX: in the previous sets dots are $3x3$ pixels, here they are $1x1$ pixel of size.
\item Hard: in this case, the set of all possible coordinate pairs of pixels is divided into two disjoint sets. Then the training images are generated from one set and the test images are generated from the other set.
The histograms of the generated train and test images are 
0: 6751,
1: 425,
2: 6651,
3: 6656,
4: 6531,
5: 6646,
6: 6678,
7: 6482,
8: 6715,
9: 6465
and
0: 1107,
1: 59,
2: 1146,
3: 1089,
4: 1045,
5: 1101,
6: 1113,
7: 1118,
8: 1141,
9: 1081
where the 22*22=484 pixels are divided into two disjoint sets of sizes 425 and 59.
\item Hard 1PX: this set is generated by the same method as the previous one, but dots are $1x1$ pixel.
\end{itemize}

The second series contains training and test images only of size $10x10$ pixels with dots of 1x1 pixel described precisely by the following:
\begin{itemize}
\item Disjunct: the same as above but there is exactly one training and exactly one test image that contain no dots.
The histograms of the generated train and test images are 
0: 1,
1: 90,
2: 4455,
3: 7926,
4: 7806,
5: 8008,
6: 7940,
7: 8069,
8: 7872,
9: 7833
and
0: 1,
1: 10,
2: 438,
3: 1382,
4: 1315,
5: 1352,
6: 1347,
7: 1441,
8: 1379,
9: 1335.
\item Hard: as in the previous case and here also the 0 dots are handled standalone.
The histograms of the generated train and test images are 
0: 1,
1: 84,
2: 3486,
3: 8126,
4: 7943,
5: 8034,
6: 8061,
7: 8115,
8: 8003,
9: 8147
and
0: 1,
1: 16,
2: 120,
3: 560,
4: 1567,
5: 1571,
6: 1518,
7: 1501,
8: 1534,
9: 1612,
where the 10*10=100 pixels are divided into two disjoint sets of sizes 84 and 16.
\item Disjunct pow 102x+, Hard pow 102x+: as in the previous ones but the probability distribution function of generating $n \in \{1, \dots, m\}$ dots in train images is the following 
\[
F(x)=
\begin{cases}
0& \text{for $x\le 1$}\\
\frac{10^{2x}}{10^{2m}}& \text{for $ 1< x \le m$}\\
1& \text{for $x\ge m$}\\
\end{cases}
\]
where $m$ denotes the maximum digit. In cases of these training and test sets, $m$  is equal to 9. The reason of choosing this distribution is that there are 
$100!/(100-n)!$ possible $10x10$ images that contain exactly $n$ (different, order matters) dots (variations without repetition).
To be more precise, in our case it is equal to $\binom{100}{n}$ because all n pixels are the same color (order does not matter, combinations without repetition)\footnote{In practice, histograms of the generated images follow the case of combination without repetition due to the uniqueness condition of the Disjunct (and all further) datasets.}.
It should be noticed that number of dots in the test images still follows uniform distribution. 
In addition, due to using the above distribution produces histograms for example like this 0: 1, 7: 5, 8: 600, 9: 59394 where the numbers of dots 2, 3, 4, 5, and 6 are missing, therefore the ten percent of generation of training images follows uniform distribution. 
In the case labelled by \enquote{Disjunct pow 102x+} the histograms of the generated train and test images are 
0: 1,
1: 71,
2: 719,
3: 717,
4: 720,
5: 770,
6: 763,
7: 792,
8: 1302,
9: 54145
and
0: 1,
1: 29,
2: 1246,
3: 1234,
4: 1222,
5: 1275,
6: 1260,
7: 1248,
8: 1204,
9: 1281.
In the other (\enquote{Hard pow 102x+}) case the histograms are 
0: 1,
1: 84,
2: 732,
3: 751,
4: 711,
5: 724,
6: 773,
7: 766,
8: 1199,
9: 54259
and
0: 1,
1: 16,
2: 120,
3: 560,
4: 1525,
5: 1617,
6: 1523,
7: 1574,
8: 1512,
9: 1552,
where the 10*10=100 pixels are divided into two disjoint sets of sizes 84 and 16.
\item From 4H-102x+ to 8H-102x+ the sets are the same as Hard pow 102x+ but $m=4, 5, 6, 7, 8$ respectively. The histograms of one of these cases labelled by \enquote{4H-102x+} can be seen in Table. \ref{table_4H102x}. 
\end{itemize}

\begin{table}[!h]
\renewcommand{\arraystretch}{1.3}
\caption{The histogram of the dataset SMNIST for Humans Series 2/4H-102x+. The total 100 pixels of an image of size 10x10 are divided into two disjoint sets where the size of one is 72 and the size of the other one is 28. The column labelled \enquote{theoretical} shows the possible number of images that contain $n=0,1,2,3,4$ dots placed on different places, while the column labelled \enquote{statistics} contains the number of generated images.}
\label{table_4H102x}
\centering
\scalebox{.75}{
\begin{tabular}{|c|r|r|r|r|}
\hline
 72/28& \multicolumn{2}{c}{theoretical}&\multicolumn{2}{|c|}{statistics}\\
\hline
\bfseries dots& \bfseries train & \bfseries test& \bfseries train & \bfseries test \\
\hline
0& 1					&1			&1 		&1\\
1& 72					&28		&72		&28\\
2& 2556				&378		&1925 	&378\\
3& 59640			&3276	&2574 	&3276\\
4& 1.02879e+06	&20475	&55428	&6317\\
\hline
\end{tabular}
}
\end{table}

All data used in this paper can be found at \url{http://smartcity.inf.unideb.hu/~norbi/SMNIST/}. 
The same data can also be found at GitLab \cite{SMNISTREPO} under the directory \texttt{Datasets/SMNIST}.

\subsection{Running results}

For measurements we have used the following well known programs and models with default or different settings and with minor modifications in some certain cases. 

\begin{itemize}
\item
Tensorflow \cite{TF}, \cite{TFREPO} 0.9.0, mnist\_softmax.py (softmax regression),
\url{https://github.com/tensorflow/tensorflow/releases/tag/v0.9.0} (running with TF version 1.13.1)
\item
Tensorflow 0.9.0, mnist\_softmax.py UDPROG is the same as the previous one but it contains extensions for printing out debug messages (for example it draws the well-known visualizations of MNIST tutorials\footnote{See, for example, \url{https://tensorflow.rstudio.com/tensorflow/articles/tutorial_mnist_beginners.html}.} shown in Fig \ref{W0-MNIST-s1}, \ref{W0-NOC-s1} and \ref{W0-H1PX-s1}).
\item
Tensorflow 1.4, mnist\_deep.py (convnet), 
\url{https://github.com/tensorflow/tensorflow/blob/r1.4/tensorflow/examples/tutorials/mnist/mnist_deep.py} (running with TF version 1.13.1).
\item
Keras \cite{KERASREPO}, mnist\_cnn.py (convnet), 
\url{https://github.com/keras-team/keras/blob/master/examples/mnist_cnn.py} (running with TF version 1.13.1).
\item
PyTorch \cite{PYT}, \cite{PYTREPO}, cifar10\_tutorial.py (convnet),
\url{https://github.com/pytorch/tutorials/blob/master/beginner_source/blitz/cifar10_tutorial.py}.
\item
deeplearning4j \cite{DL4JREPO}, based on the LeNet \cite{MNIST1} MNIST example of  
\url{https://github.com/deeplearning4j/dl4j-examples/blob/master/dl4j-examples/src/main/java/org/deeplearning4j/examples/convolution/LenetMnistExample.java}.
\item Swift \cite{SWIFTREPO},  Swift TF 2 layer MLP with softmax, \url{https://colab.research.google.com/drive/1NYzgkQAc8OZHVrr-6GOVFaYVT7WjW574} partially based on \url{https://github.com/tensorflow/swift-models}.
\item Swift TF MNIST (convnet), \url{https://github.com/tensorflow/swift-models/blob/master/MNIST}.
\item Swift TF CIFAR Keras (convnet), \url{https://github.com/tensorflow/swift-models/blob/master/CIFAR/Models.swift}.
\item Swift TF CIFAR PyTorch (convnet), \url{https://github.com/tensorflow/swift-models/blob/master/CIFAR/Models.swift}.
\item Keras/Hierarchical RNN \cite{HRNN1}, \cite{HRNN2}, mnist\_hierarchical\_rnn.py, 
 \url{https://github.com/keras-team/keras/blob/master/examples/mnist_hierarchical_rnn.py}.
\item
MXNet 1.2.1 \cite{MXNETHP},
\url{https://github.com/kovacsferencz98/SMNIST_proto/blob/master/smnist_mxnet.py},
 based on a CNN MNIST example of  
\url{ttps://www.tensorflow.org/tutorials/estimators/cnn} and Apache MXNet \url{https://mxnet.incubator.apache.org/versions/master/tutorials/python/mnist.html}.
\item Lasagne \cite{lasagne}, 
\url{https://github.com/Lasagne/Lasagne/blob/master/examples/mnist.py}, (convnet).
\end{itemize}

Table \ref{table_wellknown}, \ref{table_s2} and \ref{table_s2_n} contain the test accuracies of runs of the these investigated programs. All datasets shown is these tables in addition shown in Table \ref{table_r2s1_n} and \ref{table_r2s2_n} contain 60.000 train and 10.000 test images. Finally, it should be noted that some investigated programs are very similar to each other that also plays a validating role.

\subsection{Measurements with SMNIST for Machines Series 1}

\begin{figure}[h!]
    \centering
    \begin{subfigure}[t]{.25\linewidth}
\centering
\includegraphics[scale=.2]{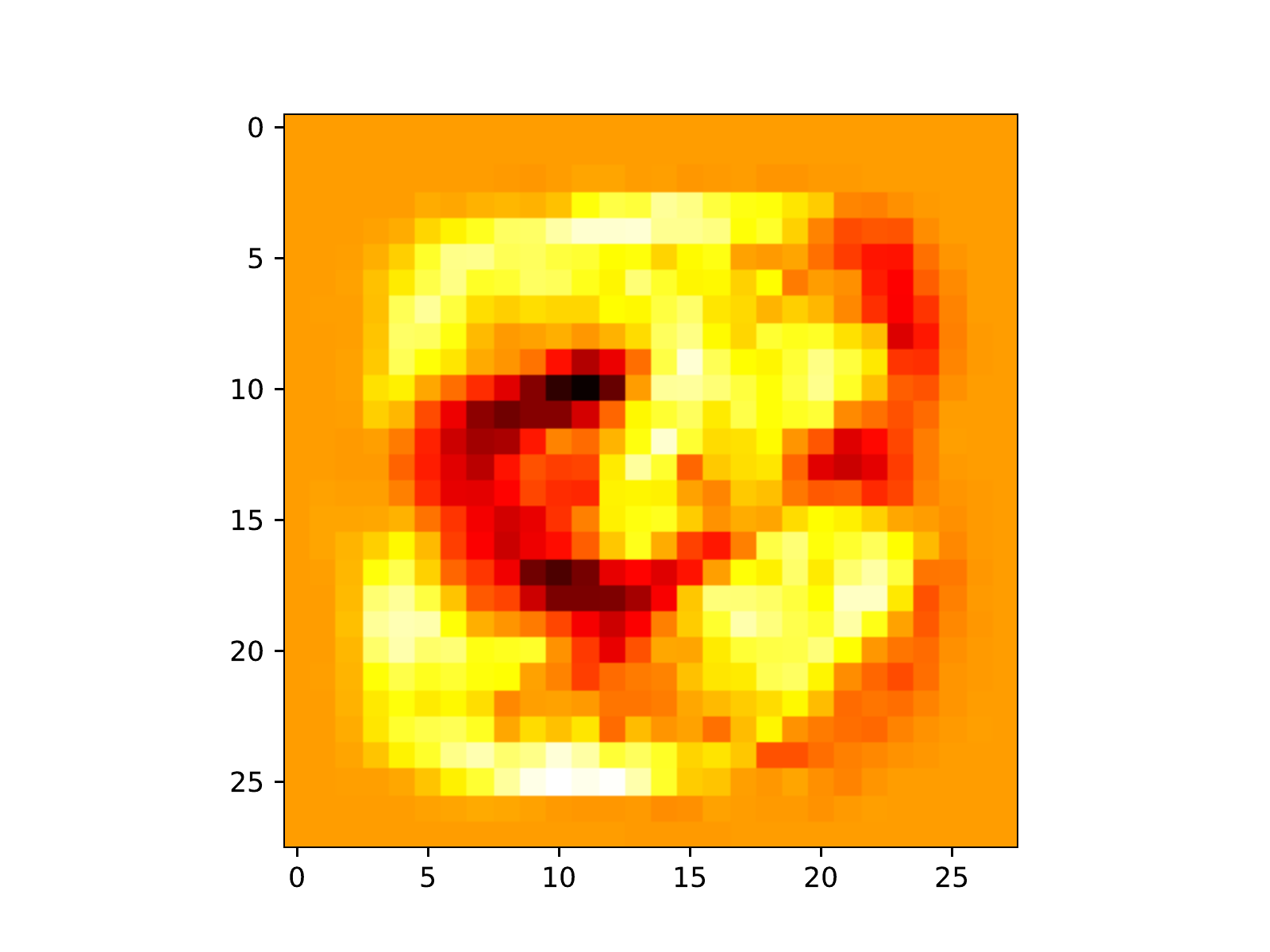}
\caption{The well known typical weights for classification of 3 in Tensorflow 0.9.0, mnist\_softmax.py (UDPROG) using the classical MNIST dataset. It may be noticed that the positive weight values draw out the silhouette of the digit 3.}
\label{W0-MNIST-s1}
    \end{subfigure}
    \hskip2em
    \begin{subfigure}[t]{.25\linewidth}
\centering
\includegraphics[scale=.2]{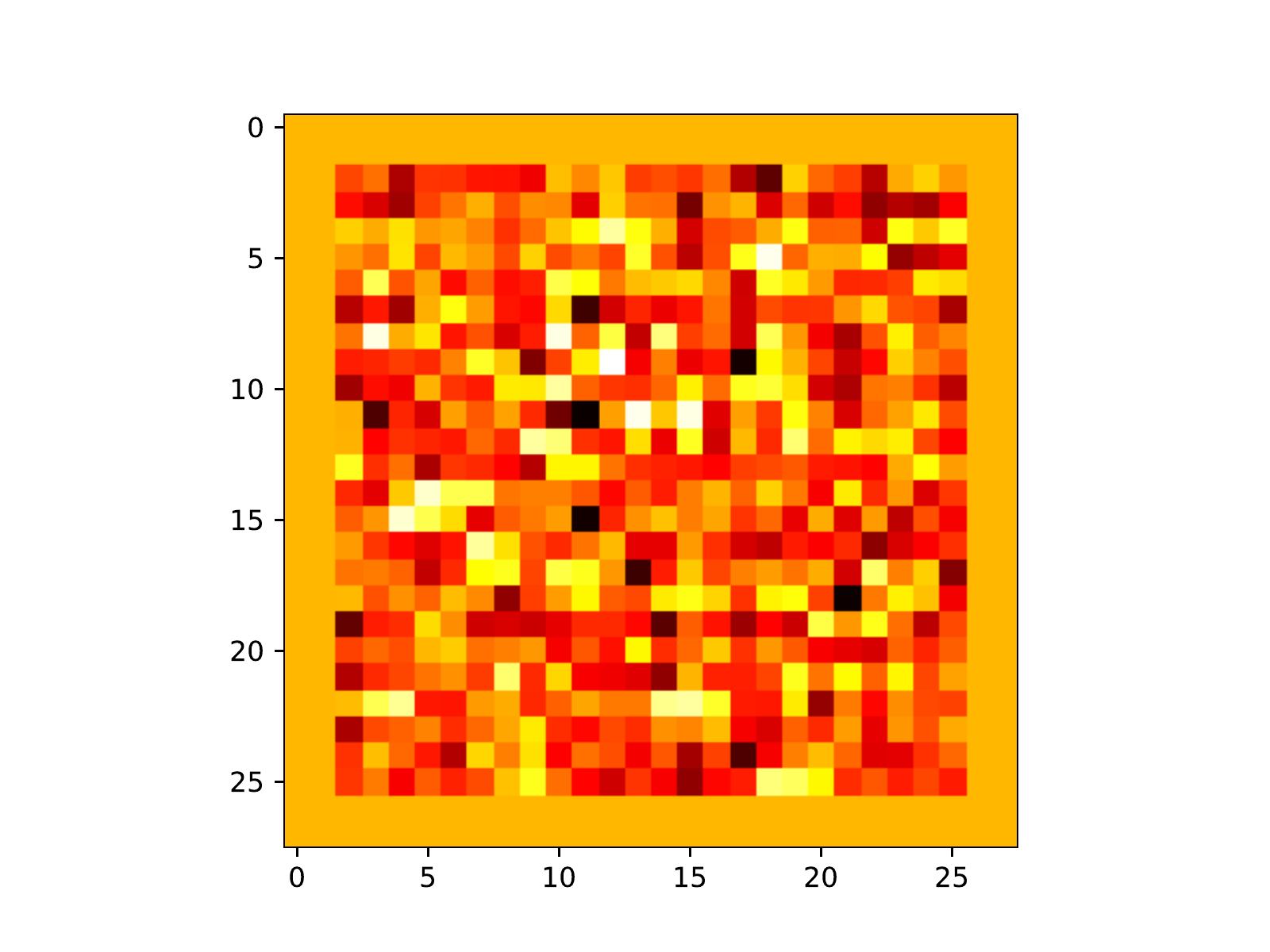}
\caption{Weights for classification of 3 in Tensorflow 0.9.0, mnist\_softmax.py (UDPROG) using the SMNIST Series 1/No-Ctrg dataset. The images of Series 1 datasets have a rectangular border of some pixels because the coordinates of dots are generated from range $[4, 24]$.}
\label{W0-NOC-s1}
    \end{subfigure}
    \hskip2em
    \begin{subfigure}[t]{.25\linewidth}
\centering
\includegraphics[scale=.2]{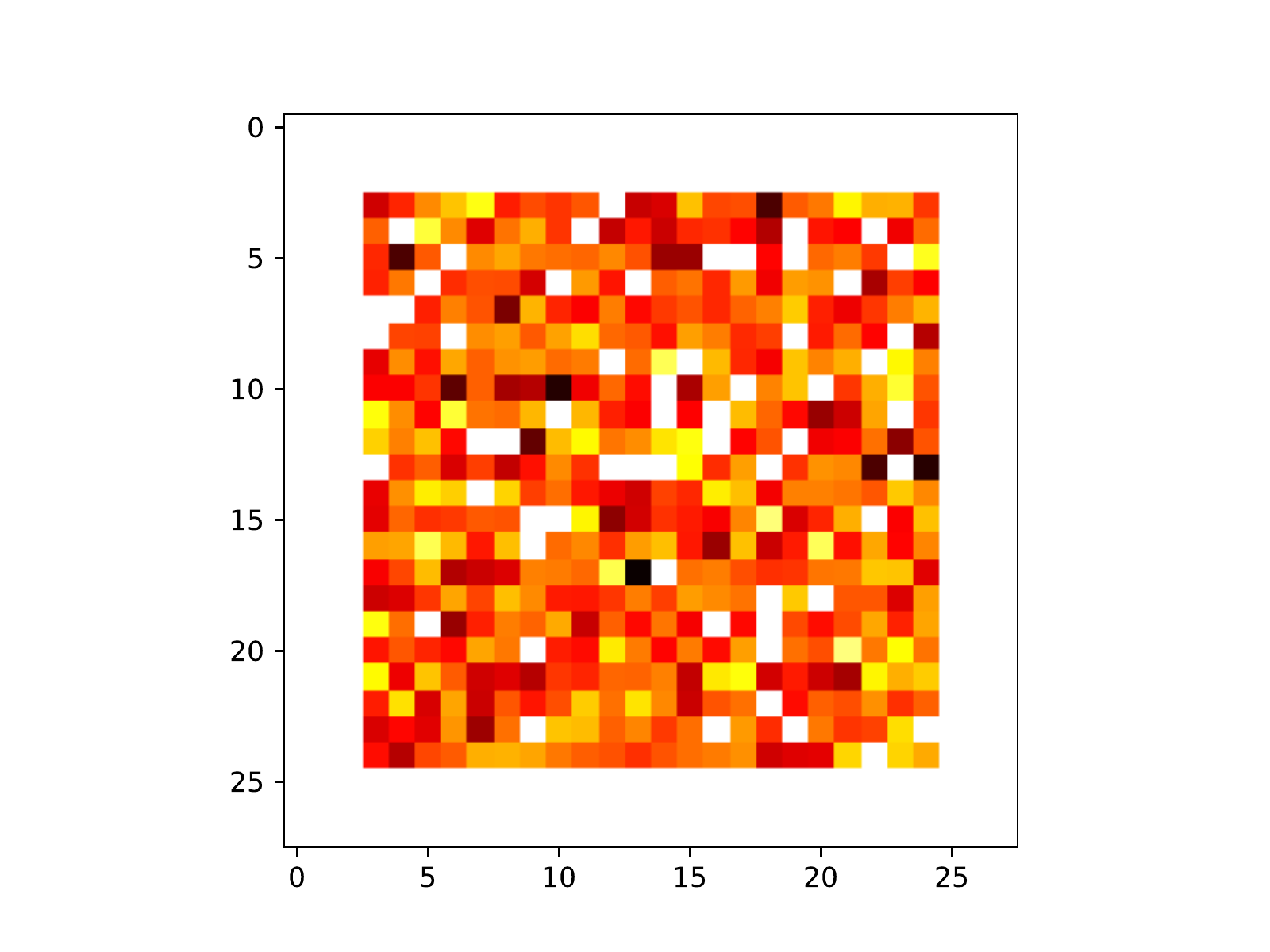}
\caption{Weights for classification of 3 in Tensorflow 0.9.0, mnist\_softmax.py  (UDPROG) using the SMNIST Series 1/H-1PX dataset. The 59 test pixels (and the 300 pixels of the border) are white.}
\label{W0-H1PX-s1}
    \end{subfigure}
    \caption{The well-known visualizations of weights of regression MNIST tutorials.}
\end{figure}

It is quite obvious that all programs produces good performance on the original MNIST dataset as it can be seen in the first column of Table \ref{table_wellknown}.
The running results for Series 1 of our datasets are shown in further columns. In first two rows, it can be seen that softmax regression models do not perform well but it is not surprising if we take a look to Fig \ref{W0-MNIST-s1}, \ref{W0-NOC-s1} and \ref{W0-H1PX-s1} where we can compare for example the visualizations of weights for classification of the digit 3. In contrast the more sophisticated models like the deep CNNs perform on the SMNIST for Machines dataset significantly better than the softmax regression.

Out of curiosity, we transferred the original PyTorch model into a DQN \cite{ATARIDQN} model and tested its performance. By transferring the model, we guaranteed that the difference between them could only originate from the different approaches (supervised learning vs. reinforcement learning). We implemented our own environment, where at every step, the model had to guess the numbers on a specific amount of images. If the correct guesses were above a certain threshold, we allowed it to continue playing, but at the same time, we increased the threshold. If the model’s performance dropped below this said threshold, the episode ended. The images were all sampled randomly from the original (Series 1/Naive) dataset. The model’s accuracy improved firmly over time, however, despite our efforts, in overall, the DQN model produced significantly worse results, such as accuracies around 0.4, 0.3 or even 0.2. We tried changing the number of episodes, the sampling procedure, and other hyperparameters such as gamma, epsilon, memory size, etc. but all to no avail.

\begin{table}[!h]
\renewcommand{\arraystretch}{1.3}
\caption{Measurements with SMNIST for Machines Series 1.}
\label{table_wellknown}
\centering
\scalebox{.75}{
\begin{tabular}{|p{3cm}|c|c|c|c|c|c|c|}
\hline
\bfseries Program& \bfseries MNIST &\bfseries Naive & \bfseries No-Ctrg  & \bfseries Disjunct& \bfseries D-1PX  & \bfseries Hard& \bfseries H-1PX\\
\hline\hline
Tensorflow 0.9.0, mnist\_softmax.py& 
0.9166&
0.6078&
0.6233&
0.5616&
0.3888& 
0.5779&
0.1107
\\
\hline
Tensorflow 0.9.0, mnist\_softmax.py, UDPROG&
0.9187&
0.6249&
0.6072&
0.5959&
0.4397& 
0.6025&
0.1107
\\
\hline
Tensorflow 1.4, mnist\_deep.py&
0.9925&
0.9787&
0.9558&
0.9608&
0.9903& 
0.9592&
0.9941
\\
\hline
Keras 2.2.4, mnist\_cnn.py&
0.9908&
0.9415&
0.9268&
0.9446&
0.9997& 
0.911&
0.9997
\\
\hline
Keras/Hierarchical RNN&
0.9858&
0.965&
0.9828&
0.9754&
0.9974& 
0.9386&
0.9655
\\
\hline
PyTorch, cifar10\_tutorial.py&
0.9907&
1.0&
0.9932&
0.8973&
0.9957& 
0.8661&
0.88\\
\hline
deeplearning4j LeNet MNIST&
0.9848&
0.9929&
0.9842&
0.9638&
0.9886&
0.9496&
0.9957
\\
\hline
MXNet 1.2.1, smnist\_mxnet.py&
0.991&
0.9717&
0.9763&
0.9436&
0.9842&
0.8911&
0.9843
\\
\hline
Lasagne, mnist.py&
0.9924&
0.9362&
0.9238&
0.9235&
0.9874&
0.8970&
0.9856
\\
\hline
\end{tabular}
}
\end{table}

\subsection{Measurements with SMNIST for Machines Series 2}

In this series we have moved from investigation of images of 28x28 size to images of 10x10 size. In addition, we manipulate the distribution of images in the train datasets. The probability of generating an image that contains n dots is roughly proportional to how many possibilities there are to place n pixels on an image of 10x10 pixels.
The precise details can be found in the previous description of datasets.
The performance of most of the tested programs has deteriorated on the data H-102x+ (see the last column of Table \ref{table_s2}). Therefore it has been splitted into further five parts (from 4H-102x+ to 8H-102x+) for further investigation that can be found in Table \ref{table_s2_n}. Based on the experiments SMNIST for Humans, we would except intuitively that performance starts to deteriorate with increasing number of dots. Actually in most tested cases this assumption is met but there also are such models where it is not true, see for example the row of Keras/Hierarchical RNN. It can intuitively be summarized that all tested programs show good performance when the number of dots does not exceed the limit of capacity of OFS measured in humans. That is the tested ANNs are supposed to be able to learn to distinguish numerosities with better accuracy when these numerosities are roughly smaller than 4. A direct experiment with smaller number of dots (H3-102x+) can be found in next section. 

\begin{table}[!h]
\renewcommand{\arraystretch}{1.3}
\caption{Measurements with SMNIST for Machines Series 2}
\label{table_s2}
\centering
\scalebox{.75}{
\begin{tabular}{|p{3cm}|c|c|c|c|}
\hline
\bfseries Program& \bfseries Disjunct &\bfseries Hard & \bfseries D-102x+  & \bfseries H-102x+ \\
\hline\hline
Tensorflow 0.9.0, mnist\_softmax.py, UDPROG&
0.6066&
0.056&
0.1281&
0.1512  
\\
\hline
Keras 2.2.4, mnist\_cnn.py&
0.8822&
0.7648&
0.8145&
0.4625\\
\hline
Keras/Hierarchical RNN&
0.9995&
0.9999&
0.9965&
0.9897\\
\hline
PyTorch, cifar10\_tutorial.py&
0.9528&
0.6243&
0.8776&
0.5365
\\
\hline
deeplearning4j LeNet MNIST&
0.8488&
0.4895&
0.2757&
0.2388
\\
\hline
MXNet 1.2.1, smnist\_mxnet.py&
0.653&
0.4013&
0.3668&
0.3005
\\
\hline
\end{tabular}
}
\end{table}

\begin{table}[!h]
\renewcommand{\arraystretch}{1.3}
\caption{Measurements with SMNIST for Machines Series 2 with particular attention to  the further breakdown of the set Hard pow 102x (H-102x+).}
\label{table_s2_n}
\centering
\scalebox{.75}{
\begin{tabular}{|p{3cm}|c|c|c|c|c|}
\hline
\bfseries Program& \bfseries 4H-102x+ &\bfseries 5H-102x+ & \bfseries 6H-102x+  & \bfseries 7H-102x+& \bfseries 8H-102x+ \\
\hline\hline
Tensorflow 0.9.0, mnist\_softmax.py& 
0.6317&
0.3188&
0.2334&
0.1936&
0.1658
\\
\hline
Keras 2.2.4, mnist\_cnn.py&
0.9099&
0.7055&
0.7599&
0.7285&
0.6568
\\
\hline
Keras/Hierarchical RNN&
0.9993&
0.9996&
0.9442&
0.9996&
0.9993
\\
\hline
PyTorch, cifar10\_tutorial.py&
0.8758&
0.8589&
0.723&
0.556&
0.6733
\\
\hline
deeplearning4j LeNet MNIST&
0.7743&
0.5329&
0.4770&
0.3671&
0.2977
\\
\hline
Swift TF MNIST
&0.6432
&0.4906
&0.3102
&0.2896
&0.1819
\\
\hline
Swift TF CIFAR PyTorch&
0.6729&
0.6218&
0.4796&
0.4156&
0.4802
\\
\hline
\end{tabular}
}
\end{table}

\section{Semantic MNIST for Anyone}

\begin{figure}[h!]
    \centering
    \begin{subfigure}{.4\linewidth}
    	\centering
        \includegraphics[scale=5.5, frame]{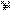}
        \caption{smnistg-train-6-7}
    \end{subfigure}
    \hskip2em
    \begin{subfigure}{.4\linewidth}
    	\centering
        \includegraphics[scale=5.5, frame]{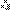}
        \caption{smnistg-train-6-8}
    \end{subfigure}
    \caption{SMNIST for Anyone, Series 1. Both images contain exactly 6 'X's.\label{r2s1}}
\end{figure}

\begin{figure}[h!]
    \centering
    \begin{subfigure}{.4\linewidth}
    	\centering
        \includegraphics[scale=5.5, frame]{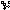}
        \caption{SSOS+O, $\text{S}^3\text{O}^2+$}
    \end{subfigure}
    \hskip .5cm
    \begin{subfigure}{.4\linewidth}
    	\centering
        \includegraphics[scale=5.5, frame]{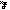}
        \caption{SXXO+X, $\text{X}^3SO+$}
    \end{subfigure}\\
    \begin{subfigure}{.4\linewidth}
    	\centering
        \includegraphics[scale=5.5, frame]{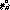}
        \caption{SSOOXXXSX, $\text{X}^4\text{S}^3\text{O}^2$}
    \end{subfigure}
    \hskip .5cm    
    \begin{subfigure}{.4\linewidth}
    	\centering
        \includegraphics[scale=5.5, frame]{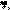}
        \caption{SXXO+X++S, $\text{X}^3\text{+}^3\text{S}^2O$}
    \end{subfigure}    \caption{SMNIST for Anyone, Series 2. Images contain exactly 6 or 9 symbols of the following: 'X', 'O', '+' and square outline ('S').\label{r2s2}}
\end{figure}

The SMNIST for Anyone is a natural further development of SMNIST for Machines.
Machines can perform this test so do humans. But at this moment we have no test filling program for humans (technically, it will be based on the previously presented SMNIST for Humans Android application).
The SMNIST for Anyone datasets are organized into two series. They are the same as the 4H-102x+, \dots, 9H-102x+(=H-102x+) datasets of the previous section where dots are replaced by 3x3 pixels patterns of the objects 'X', 'O', '+' and square outline ('S') as it can be seen in Fig. \ref{r2s1} and  \ref{r2s2}. 
It is important to highlight that this test is not uniquely determined because 
in many cases it is not clear how many objects have been placed on the images.

\subsection{Measurements with SMNIST for Anyone Series 1}

The images of Series 1 contain only 3x3 pixels binary patterns of 'X's. In all cases the performance has already been deteriorated with increasing number of dots as it can be seen in Table \ref{table_r2s1_n}.

\begin{table}[!h]
\renewcommand{\arraystretch}{1.3}
\caption{Measurements with SMNIST for Anyone Series 1.}
\label{table_r2s1_n}
\centering
\scalebox{.75}{
\begin{tabular}{|p{3cm}|c|c|c|c|c|c|}
\hline
\bfseries Program& \bfseries H4-102x+ &\bfseries H5-102x+ & \bfseries H6-102x+  & \bfseries H7-102x+& \bfseries H8-102x+ & \bfseries H9-102x+\\
\hline\hline
Tensorflow 0.9.0, mnist\_softmax.py&
0.6317&
0.3188&
0.2334&
0.1942&
0.1671&
0.1402
\\
\hline
Keras 2.2.4, mnist\_cnn.py&
0.836&
0.7546&
0.6914&
0.6702&
0.6233&
0.5913
\\
\hline
Keras/Hierarchical RNN&
0.8498&
0.7152&
0.6896&
0.6537&
0.5144&
0.5498
\\
\hline
deeplearning4j LeNet MNIST&
0.6862&
0.6764&
0.3845&
0.3394&
0.3008&
0.2245
\\
\hline
\end{tabular}
}
\end{table}

\subsection{Measurements with SMNIST for Anyone Series 2}

The images of Series 2 may contain any of the symbols 'X', 'O', '+' and square outline ('S'). As shown in Table \ref{table_r2s2_n} we experience the same performance as observed in the previous series of experiments.

\begin{table}[!h]
\renewcommand{\arraystretch}{1.3}
\caption{Measurements with SMNIST for Anyone Series 2.}
\label{table_r2s2_n}
\centering
\scalebox{.75}{
\begin{tabular}{|p{3cm}|c|c|c|c|c|c|}
\hline
\bfseries Program& \bfseries H4-102x+ &\bfseries H5-102x+ & \bfseries H6-102x+  & \bfseries H7-102x+& \bfseries H8-102x+ & \bfseries H9-102x+\\
\hline\hline
Tensorflow 0.9.0, mnist\_softmax.py&
0.6317&
0.3217&
0.2388&
0.192&
0.1638&
0.1399
\\
\hline
Keras 2.2.4, mnist\_cnn.py&
0.7959&
0.5627&
0.6036&
0.5504&
0.516&
0.4696
\\
\hline
Keras/Hierarchical RNN&
0.7366&
0.595&
0.5198&
0.5014&
0.5099&
0.4406
\\
\hline
deeplearning4j LeNet MNIST&
0.6558&
0.4166&
0.3566&
0.3203&
0.2743&
0.2304
\\
\hline
\end{tabular}
}
\end{table}

Finally, we have also made a direct experiment with maximum 3 number of objects. The properties of its dataset called H3-102x+ can be seen in Table \ref{table_H3102x}. As we expected according to our thesis, the tested programs perform well in this experiment, for example the \enquote{Keras 2.2.4, mnist\_cnn.py} produces  an accuracy of 0.9436 or \enquote{Keras/Hierarchical RNN} produces an accuracy of 0.9522.

\begin{table}[!h]
\renewcommand{\arraystretch}{1.3}
\caption{The histogram of the dataset Series 2/H3-102x+. This contains 10.000 test and only 30.000 train images. The 10*10=100 pixels are divided into two disjoint sets of sizes 57 and 43. The column labelled \enquote{theoretical} shows the possible number of images that contain $n=0,1,2,3$ different dots, while the column labelled \enquote{statistics} contains the number of generated images.}
\label{table_H3102x}
\centering
\scalebox{.75}{
\begin{tabular}{|c|r|r|r|r|}
\hline
57/43& \multicolumn{2}{c}{theoretical}&\multicolumn{2}{|c|}{statistics}\\
\hline
\bfseries dots& \bfseries train & \bfseries test& \bfseries train & \bfseries test \\
\hline
0& 1					&1			&1 		&1\\
1& 57					&43		&57		&43\\
2& 1596				&903		&1567 	&903\\
3& 29260			&12341	&28375 	&9053\\
\hline
\end{tabular}
}
\end{table}

\section{Conclusion and Further Work}

In all software experiments of this study we investigate the numerosity of quantities. 
The SMNIST for Humans experimental results are well in accordance with observations from cognitive psychology literature.  
Based on our SMNIST for Humans and SMNIST for Anyone experiences we can intuitively formulate our thesis as follows: image classification (such as MNIST or CIFAR-10) ANNs can learn to distinguish numerosities with better accuracy when these numerosities are smaller than the capacity of OFS measured in humans (that is roughly smaller than 4).

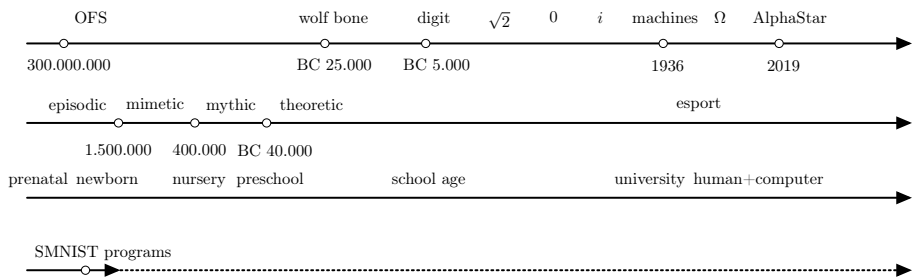
\begin{figure*}[!h]
\centering\scalebox{.62}{\input{omega.tex}}\caption{This Haeckel-like figure contains four timelines. Intuitively, the first one tries to outline a mental evolutionary phylogeny of humans especially focusing to the notion of number \cite{Klix}. The second one shows Merlin Donald's distinguishment between stages of mental evolution \cite{MerlinDonald}. The third one presents the stages of ontogeny of the notion of number in humans. And finally, the last timeline tries to introduce ontogeny of the notion of number in machines. It should be noted that timelines are not linear: the concrete dates do not matter, but their order does. \label{fig_omega}}
\end{figure*}

Fig. \ref{fig_omega} outlines a conceptual framework for analyzing the notion of number in humans and machines. The first timeline of the figure tells that OFS may have begun evolving 300 million years ago \cite{Number}. This is followed by the \enquote{wolf bone}, an assumed tally based external counting device \cite{Klix}. Then appears the first numerical digit \cite{Klix}. The $\sqrt 2$ had already been approximated in sexagesimal arithmetic by ancient Babylonians \cite{sqrt2}. It stands for the appearance of the notion of numeral systems. The imaginary $i$ simply denotes the extension of the notion of number with the complex numbers. Turing's famous study \cite{Turing1936} of 1936 indicates the onset of digital computers. There are numbers such as $\Omega$ \cite{Omega}  simply would not exist without digital computers. Finally, the breakthrough machine learning application called AlphaStar \cite{AlphaStarB} represents today's computer programs. It is important to emphasize that all mentioned devices including software are based on Donaldian external storage systems of theoretical culture as it is expressed by the second timeline of the figure. Here we expand Donald's three stages of mental evolution  \cite{MerlinDonald} with an additional stage called \enquote{esport}. The expansion of stages is not rare in the literature, see for example \cite{MentEvolProgs} where some timeline dates are little bit different from ours, but similarly, the additional stage is focused on computer programs. The esport and computer gaming like the computer programs in general have also been implemented in the Donaldian external memory of theoretical culture. Our utopian goal is to create an open source esport game that would be able to function as Leibniz's \enquote{characteristica universalis} and as such can express some notion of number \cite{MITEL}.

The stages of ontogeny of the notion of number in humans is presented on third timeline. By the time children reached school age they had acquired the language. 
Preschool children have already played electronic games but the minimum age limit of participation in esport tournaments is various in the range from 12 to 18 years old. Moreover our utopian interpretation of the new stage labelled by esport is shifted to university years because here we are thinking of the aforementioned esport game to be developed.

The digital computers are products of pure theoretical culture. For example, in this sense, the \enquote{AI winter} can be interpreted as the time required for machines of theoretical culture to learn to work on different lower Donaldian layer of culture \cite{MITEL}. In this interpretation, it may be a possible solution for the Moravec's paradox. And it is also clear that it does not work backwards, for example just think of antinomies of naive set theory, where the source of problems is that we have tried to handle entities of theoretical culture at a lower layer of culture specifically with natural spoken (or written) language tools that are specific for the Donaldian mythical culture. From this point of view, the SMNIST programs that are entities of theoretical culture for today seems can solve the problem of subitizing that is part of the episodic culture. It is presented on fourth timeline. 
Why, for example, the AlphaStar is not shown on right side of this timeline? Due to computer programs of today are using the man's notion of number rather than their own one because it does not yet exist.

\section{Acknowledgment}

The authors thank the students of the BSc course titled \enquote{High Level Programming Languages} at the University of Debrecen and the members of the UDPROG Facebook community \url{https://www.facebook.
com/groups/udprog} for performing the SMNIST for Humans test.
In addition, the authors thank Krisztina Gy\H{o}ri for the proofreading of the manuscript. 

This work was supported by the construction EFOP-3.6.3-VEKOP-16-2017-00002. The project was supported by the European Union, co-financed by the European Social Fund.

Author contributions were the following: N. B. conceived the idea of SMNIST for Machines, SMNIST for Humans and SMNIST for Anyone, developed the generator and Android programs, collected the SMNIST for Humans data from the UDPROG community and analyzed the measurements. D. P. provided cognitive psychological background. N. B., G. B., M. Sz., M. B., G. Sz., F. K., E. Sz. V. performed SMNIST for Machines computations. 
V. Sz. S. and  L. K.  performed some computations with a previous version of SMNIST for Machines datasets.
All authors wrote the paper and discussed the results.

\bibliographystyle{alpha}
\bibliography{smnist}

\end{document}

%% file: omega.tex
\begin{tikzpicture}[line cap=round,line join=round,>=triangle 45,x=1cm,y=1cm]
\draw [->,line width=1.2pt] (-4.28,-2.8) -- (14.5,-2.8);
\draw (-4.4,-3.02) node[anchor=north west] {300.000.000};
\draw (-3.4,-2) node[anchor=north west] {OFS};
\draw (1.36,-2) node[anchor=north west] {wolf bone};
\draw (5.38,-2) node[anchor=north west] {$\sqrt{2}$};
\draw (6.68,-2) node[anchor=north west] {$0$};
\draw (7.7,-2) node[anchor=north west] {$i$};
\draw (-3.94,-3.85) node[anchor=north west] {episodic};
\draw (-2.3,-3.85) node[anchor=north west] {mimetic};
\draw (-0.6,-3.85) node[anchor=north west] {mythic};
\draw (0.96,-3.85) node[anchor=north west] {theoretic};
\draw (9.38,-3.85) node[anchor=north west] {esport};
\draw [->,line width=1.2pt] (-4.28,-4.5) -- (14.5,-4.5);
\draw (0.06,-4.84) node[anchor=north west] {BC 40.000};
\draw (-1.32,-4.84) node[anchor=north west] {400.000};
\draw (-3.18,-4.84) node[anchor=north west] {1.500.000};
\draw [->,line width=1.2pt] (-4.28,-6.1) -- (14.5,-6.1);
\draw (8.84,-3.04) node[anchor=north west] {1936};
\draw (-4.8,-5.45) node[anchor=north west] {prenatal};
\draw (-3.36,-5.45) node[anchor=north west] {newborn};
\draw (-1.32,-5.54) node[anchor=north west] {nursery};
\draw (0.04,-5.45) node[anchor=north west] {preschool};
\draw (3.34,-5.45) node[anchor=north west] {school age};
\draw (8.06,-5.45) node[anchor=north west] {university};
\draw (9.74,-5.45) node[anchor=north west] {human+computer};
\draw [->,line width=1.2pt] (-4.28,-7.65) -- (-2.3,-7.65);
\draw (-4.24,-7.0) node[anchor=north west] {SMNIST programs};
\draw (1.32,-3) node[anchor=north west] {BC 25.000};
\draw (3.58,-3) node[anchor=north west] {BC 5.000};
\draw (10.18,-2) node[anchor=north west] {$\Omega$};
\draw (11,-2) node[anchor=north west] {AlphaStar};
\draw (11.3,-3.04) node[anchor=north west] {2019};
\draw (3.88,-2) node[anchor=north west] {digit};
\draw (8.44,-2) node[anchor=north west] {machines};
\draw [->,line width=1.2pt,dotted] (-2.3,-7.65) -- (14.5,-7.65);
\begin{scriptsize}
\draw [fill=feher] (-3.5,-2.8) circle (2.5pt);
\draw [fill=feher] (2.0399864563431693,-2.8) circle (2.5pt);
\draw [fill=feher] (4.180156686684751,-2.8) circle (2.5pt);
\draw [fill=feher] (-0.7198977506036259,-4.5) circle (2.5pt);
\draw [fill=feher] (0.8037863832389611,-4.5) circle (2.5pt);
\draw [fill=feher] (-2.3399384290138023,-4.5) circle (2.5pt);
\draw [fill=feher] (9.220096439194165,-2.8) circle (2.5pt);
\draw [fill=feher] (-3.039944246036844,-7.65) circle (2.5pt);
\draw [fill=feher] (11.679882509713078,-2.8) circle (2.5pt);
\end{scriptsize}
\end{tikzpicture}

%% file: smnist.bbl
\newcommand{\etalchar}[1]{$^{#1}$}
\begin{thebibliography}{HOBB{\etalchar{+}}17}

\bibitem[AAB{\etalchar{+}}16]{TF}
Mart{\'{\i}}n Abadi, Ashish Agarwal, Paul Barham, Eugene Brevdo, Zhifeng Chen,
  Craig Citro, Gregory~S. Corrado, Andy Davis, Jeffrey Dean, Matthieu Devin,
  Sanjay Ghemawat, Ian~J. Goodfellow, Andrew Harp, Geoffrey Irving, Michael
  Isard, Yangqing Jia, Rafal J{\'{o}}zefowicz, Lukasz Kaiser, Manjunath Kudlur,
  Josh Levenberg, Dan Man{\'{e}}, Rajat Monga, Sherry Moore, Derek~Gordon
  Murray, Chris Olah, Mike Schuster, Jonathon Shlens, Benoit Steiner, Ilya
  Sutskever, Kunal Talwar, Paul~A. Tucker, Vincent Vanhoucke, Vijay Vasudevan,
  Fernanda~B. Vi{\'{e}}gas, Oriol Vinyals, Pete Warden, Martin Wattenberg,
  Martin Wicke, Yuan Yu, and Xiaoqiang Zheng.
\newblock Tensorflow: Large-scale machine learning on heterogeneous distributed
  systems.
\newblock {\em CoRR}, abs/1603.04467, 2016.

\bibitem[ACT19]{AlphaStar}
Kai Arulkumaran, Antoine Cully, and Julian Togelius.
\newblock Alphastar: An evolutionary computation perspective.
\newblock {\em CoRR}, abs/1902.01724, 2019.

\bibitem[Apa19]{MXNETHP}
{MXNet: A Scalable Deep Learning Framework}.
\newblock \url{https://mxnet.apache.org/}, 2019.

\bibitem[B{\'a}t19a]{MITEL}
Norbert B{\'a}tfai.
\newblock A cognitive evolutionary interpretation of artificial intelligence
  (\textit{A mesters\'eges intelligencia kognit\'iv evol\'uci\'os
  \'ertelmez\'ese}).
\newblock {Unpublished Manuscript, Original paper in Hungarian}, 2019.

\bibitem[B{\'a}t19b]{SMNISTREPO}
Norbert B{\'a}tfai.
\newblock Smnist.
\newblock \url{https://gitlab.com/nbatfai/smnist}, 2019.

\bibitem[BB89]{Chimp}
Sarah Boysen and Gary Berntson.
\newblock Numerical competence in a chimpanzee (pan troglodytes ).
\newblock {\em Journal of comparative psychology}, 103:23--31, 1989.

\bibitem[BLT{\etalchar{+}}16]{DeepMindLab}
Charles Beattie, Joel~Z. Leibo, Denis Teplyashin, Tom Ward, Marcus Wainwright,
  Heinrich K{\"{u}}ttler, Andrew Lefrancq, Simon Green, V{\'{\i}}ctor
  Vald{\'{e}}s, Amir Sadik, Julian Schrittwieser, Keith Anderson, Sarah York,
  Max Cant, Adam Cain, Adrian Bolton, Stephen Gaffney, Helen King, Demis
  Hassabis, Shane Legg, and Stig Petersen.
\newblock Deepmind lab.
\newblock {\em CoRR}, abs/1612.03801, 2016.

\bibitem[C{\etalchar{+}}19]{KERASREPO}
François Chollet et~al.
\newblock Github - keras-team/keras: Deep learning for humans.
\newblock \url{https://github.com/keras-team/keras}, 2019.

\bibitem[Cas16]{BBAI}
D.~Castelvecchi.
\newblock Can we open the black box of {AI}?
\newblock {\em Nature}, 538:20--23, 2016.

\bibitem[CBM{\etalchar{+}}19]{PYTREPO}
Ronan Collobert, Samy Bengio, Johnny Mariethoz, Adam Paszke, Soumith Chintala,
  Koray Kavukcuoglu, Clement Farabet, Leon Bottou, Iain Melvin, Jason Weston,
  et~al.
\newblock {GitHub - pytorch/pytorch: Tensors and Dynamic neural networks in
  Python with strong GPU acceleration}.
\newblock \url{https://github.com/pytorch/pytorch}, 2019.

\bibitem[{Cha}04]{Omega}
G.~J. {Chaitin}.
\newblock {{Meta Math! The Quest for Omega}}.
\newblock {\em arXiv Mathematics e-prints}, 2004.

\bibitem[DC93]{NModel}
Stanislas Dehaene and Jean-Pierre Changeux.
\newblock Development of elementary numerical abilities: A neuronal model.
\newblock {\em Journal of cognitive neuroscience}, 5:390--407, 1993.

\bibitem[DDLC98]{BrainNumbers}
Stanislas Dehaene, Ghislaine Dehaene-Lambertz, and Laurent Cohen.
\newblock Abstract representations numbers in the animal and human brain.
\newblock {\em Trends in neurosciences}, 21:355--61, 1998.

\bibitem[DM82]{Davis82}
Hank Davis and John Memmott.
\newblock Counting behavior in animals: A critical evaluation.
\newblock {\em Psychological Bulletin}, 92:547--571, 1982.

\bibitem[DN16]{OFS}
Helen~M Ditz and Andreas Nieder.
\newblock Numerosity representations in crows obey the weber-fechner law.
\newblock {\em Proc. Biol. Sci.}, 283(1827):1-- 9, 2016.

\bibitem[Don91]{MerlinDonald}
Merlin Donald.
\newblock {\em Origins of the modern mind : three stages in the evolution of
  culture and cognition}.
\newblock Harvard University Press Cambridge, Mass, 1991.

\bibitem[DP88]{Davis1988}
Hank Davis and Rachelle Pérusse.
\newblock Numerical competence in animals: Definitional issues, current
  evidence, and a new research agenda.
\newblock {\em Behavioral and Brain Sciences}, 11:561--579, 1988.

\bibitem[DSR{\etalchar{+}}15]{lasagne}
Sander Dieleman, Jan Schlüter, Colin Raffel, Eben Olson, Søren~Kaae
  Sønderby, Daniel Nouri, Daniel Maturana, Martin Thoma, Eric Battenberg, Jack
  Kelly, Jeffrey~De Fauw, Michael Heilman, Diogo~Moitinho de~Almeida, Brian
  McFee, Hendrik Weideman, Gábor Takács, Peter de~Rivaz, Jon Crall, Gregory
  Sanders, Kashif Rasul, Cong Liu, Geoffrey French, and Jonas Degrave.
\newblock Lasagne: First release., 2015.

\bibitem[DVS08]{Insect}
Marie Dacke and Mandyam V~Srinivasan.
\newblock Evidence for counting in insects.
\newblock {\em Animal cognition}, 11:683--9, 2008.

\bibitem[DWW15]{HRNN2}
Yong Du, Wei Wang, and Liang Wang.
\newblock Hierarchical recurrent neural network for skeleton based action
  recognition.
\newblock In {\em CVPR}, pages 1110--1118. IEEE Computer Society, 2015.

\bibitem[FBAH66]{size}
Gustav~Theodor Fechner, 1886-1968 Boring, Edwin~Garrigues, Helmut~E Adler, and
  Davis~H Howes.
\newblock {\em Elements of psychophysics}.
\newblock New York: Holt, Rinehart and Winston, 1966.

\bibitem[FCH02]{PIS2}
Lisa Feigenson, Susan Carey, and Marc Hauser.
\newblock The representations underlying infants' choice of more: Object files
  versus analog magnitudes.
\newblock {\em Psychological Science}, 13(2):150--156, 2002.

\bibitem[FDS04]{PISCoreSys2}
Lisa Feigenson, Stanislas Dehaene, and Elizabeth Spelke.
\newblock Core systems of number.
\newblock {\em Trends in Cognitive Sciences}, 8(7):307 -- 314, 2004.

\bibitem[FR98]{sqrt2}
David Fowler and Eleanor Robson.
\newblock Square root approximations in old babylonian mathematics: Ybc 7289 in
  context.
\newblock {\em Historia Mathematica}, 25(4):366--378, 1998.

\bibitem[Gea95]{PIS00}
David Geary.
\newblock Reflections of evolution and culture in children's cognition:
  Implications for mathematical development and instruction.
\newblock {\em The American psychologist}, 50:24--37, 1995.

\bibitem[Gea00]{PIS0}
D.C. Geary.
\newblock From infancy to adulthood: the development of numerical abilities.
\newblock {\em European Child \& Adolescent Psychiatry}, 9:11--16, 2000.

\bibitem[HOBB{\etalchar{+}}17]{Goliath}
Jose Hernandez-Orallo, Marco Baroni, Jordi Bieger, Nader Chmait, David L.~Dowe,
  Katja Hofmann, Fernando Plumed, Claes Strannegård, and Kristinn Thórisson.
\newblock A new ai evaluation cosmos: Ready to play the game?
\newblock {\em AI Magazine}, 38:66, 2017.

\bibitem[Hyd11]{PIS1}
Daniel Hyde.
\newblock Two systems of non-symbolic numerical cognition.
\newblock {\em Frontiers in human neuroscience}, 5:150--158, 2011.

\bibitem[JCD{\etalchar{+}}18]{Doom}
Max Jaderberg, Wojciech~M. Czarnecki, Iain Dunning, Luke Marris, Guy Lever,
  Antonio~Garc{\'{\i}}a Casta{\~{n}}eda, Charles Beattie, Neil~C. Rabinowitz,
  Ari~S. Morcos, Avraham Ruderman, Nicolas Sonnerat, Tim Green, Louise Deason,
  Joel~Z. Leibo, David Silver, Demis Hassabis, Koray Kavukcuoglu, and Thore
  Graepel.
\newblock Human-level performance in first-person multiplayer games with
  population-based deep reinforcement learning.
\newblock {\em CoRR}, abs/1807.01281, 2018.

\bibitem[JD02]{MentEvolProgs}
Kaput J.J. and Shaffer D.W.
\newblock On the development of human representational competence from an
  evolutionary point of view.
\newblock In Koeno Gravemeijer, Richard Lehrer, Bert~Van Oers, and Lieven
  Verschaffel, editors, {\em Symbolizing, Modeling and Tool Use in Mathematics
  Education}, volume~30 of {\em Mathematics Education Library}, pages 277--293.
  Springer, 2002.

\bibitem[JDO{\etalchar{+}}17]{LamEvol}
Max Jaderberg, Valentin Dalibard, Simon Osindero, Wojciech~M. Czarnecki, Jeff
  Donahue, Ali Razavi, Oriol Vinyals, Tim Green, Iain Dunning, Karen Simonyan,
  Chrisantha Fernando, and Koray Kavukcuoglu.
\newblock Population based training of neural networks.
\newblock {\em CoRR}, abs/1711.09846, 2017.

\bibitem[Kah74]{DataFlow}
Gilles Kahn.
\newblock The semantics of simple language for parallel programming.
\newblock In {\em IFIP Congress}, pages 471--475, 1974.

\bibitem[Kli85]{Klix}
Friedhart Klix.
\newblock {\em Az \'ebred{\H o} gondolkod\'as}.
\newblock Gondolat, Budapest, 1985.

\bibitem[Knu97]{Knuth}
Donald~E. Knuth.
\newblock {\em The Art of Computer Programming, Volume 2 (3rd Ed.):
  Seminumerical Algorithms}.
\newblock Addison-Wesley Longman Publishing Co., Inc., Boston, MA, USA, 1997.

\bibitem[Kri09]{cifar10}
A.~Krizhevsky.
\newblock {Learning Multiple Layers of Features from Tiny Images}.
\newblock Technical report, University of Toronto, Toronto, 2009.

\bibitem[LBBH98]{MNIST1}
Yann LeCun, Léon Bottou, Yoshua Bengio, and Patrick Haffner.
\newblock Gradient-based learning applied to document recognition.
\newblock In {\em Proceedings of the IEEE}, volume~86, pages 2278--2324, 1998.

\bibitem[LBD{\etalchar{+}}89]{LeCunConvNet}
Y.~{LeCun}, B.~{Boser}, J.~S. {Denker}, D.~{Henderson}, R.~E. {Howard},
  W.~{Hubbard}, and L.~D. {Jackel}.
\newblock Backpropagation applied to handwritten zip code recognition.
\newblock {\em Neural Computation}, 1(4):541--551, 1989.

\bibitem[LCB]{MNISTHP}
Yann LeCun, Corinna Cortes, and Christopher J.~C. Burges.
\newblock The mnist database of handwritten digits.
\newblock \url{http://yann.lecun.com/exdb/mnist/}.
\newblock Accessed: 2019-04-26.

\bibitem[LLJ15]{HRNN1}
Jiwei Li, Minh{-}Thang Luong, and Dan Jurafsky.
\newblock A hierarchical neural autoencoder for paragraphs and documents.
\newblock {\em CoRR}, abs/1506.01057, 2015.

\bibitem[MDM80]{dist}
Edward M.~Duncan and Carl Mcfarland.
\newblock Isolating the effects of symbolic distance, and semantic congruity in
  comparative judgments: An additive-factors analysis.
\newblock {\em Memory \& cognition}, 8:612--22, 1980.

\bibitem[MKS{\etalchar{+}}13]{ATARIDQN}
Volodymyr Mnih, Koray Kavukcuoglu, David Silver, Alex Graves, Ioannis
  Antonoglou, Daan Wierstra, and Martin Riedmiller.
\newblock Playing atari with deep reinforcement learning.
\newblock {\em eprint arXiv:1312.5602}, 2013.

\bibitem[MKS{\etalchar{+}}15]{ATARI}
Volodymyr Mnih, Koray Kavukcuoglu, David Silver, Andrei~A. Rusu, Joel Veness,
  Marc~G. Bellemare, Alex Graves, Martin~A. Riedmiller, Andreas Fidjeland,
  Georg Ostrovski, Stig Petersen, Charles Beattie, Amir Sadik, Ioannis
  Antonoglou, Helen King, Dharshan Kumaran, Daan Wierstra, Shane Legg, and
  Demis Hassabis.
\newblock Human-level control through deep reinforcement learning.
\newblock {\em Nature}, 518(7540):529--533, 2015.

\bibitem[MKTD16]{MALMO}
Johnson M., Hofmann K., Hutton T., and Bignell D.
\newblock The malmo platform for artificial intelligence experimentation.
\newblock In {\em Proc. 25th International Joint Conference on Artificial
  Intelligence}, pages 4246--4247, 2016.

\bibitem[MP43]{Neuron}
W.~S. McCulloch and W.~Pitts.
\newblock A logical calculus of the ideas immanent in nervous activity.
\newblock {\em Bulletin of Mathematical Biophysics}, 5:115--133, 1943.

\bibitem[Neu58]{Neumann}
John~von Neumann.
\newblock {\em The Computer and the Brain}.
\newblock Yale University Press, 1958.

\bibitem[Nie16]{Number}
Andreas Nieder.
\newblock The neuronal code for number.
\newblock {\em Nature reviews. Neuroscience}, 17, 2016.

\bibitem[Pep10]{Parrot}
Irene Pepperberg.
\newblock Evidence for conceptual quantitative abilities in the african grey
  parrot: Labeling of cardinal sets.
\newblock {\em Ethology}, 75:37--61, 2010.

\bibitem[PGC{\etalchar{+}}17]{PYT}
Adam Paszke, Sam Gross, Soumith Chintala, Gregory Chanan, Edward Yang, Zachary
  DeVito, Zeming Lin, Alban Desmaison, Luca Antiga, and Adam Lerer.
\newblock Automatic differentiation in {PyTorch}.
\newblock In {\em NIPS Autodiff Workshop}, 2017.

\bibitem[Pia10]{Piazza3}
Manuela Piazza.
\newblock Neurocognitive start-up tools for symbolic number representations.
\newblock {\em Trends in cognitive sciences}, 14:542--51, 11 2010.

\bibitem[PIP{\etalchar{+}}04]{Piazza1}
Manuela Piazza, Veronique Izard, Philippe Pinel, Denis Le~Bihan, and Stanislas
  Dehaene.
\newblock Tuning curves for approximate numerosity in the human intraparietal
  sulcus.
\newblock {\em Neuron}, 44:547--55, 2004.

\bibitem[PPLBD07]{Piazza2}
Manuela Piazza, Philippe Pinel, Denis Le~Bihan, and Stanislas Dehaene.
\newblock A magnitude code common to numerosities and number symbols in human
  intraparietal cortex.
\newblock {\em Neuron}, 53:293--305, 2007.

\bibitem[RKRC70]{Cat}
Thompson RF, Mayers KS, Robertson RT, and Patterson CJ.
\newblock Number coding in association cortex of the cat.
\newblock {\em Science}, 168(3928):271--273, 1970.

\bibitem[Ros58]{Perceptron}
Frank Rosenblatt.
\newblock The perceptron: A probabilistic model for information storage and
  organization in the brain.
\newblock {\em Psychological Review}, 65(6):386--408, 1958.

\bibitem[SD83]{Child1}
Antell SE and Keating DP.
\newblock Perception of numerical invariance in neonates.
\newblock {\em Child Dev.}, 54:695--701, 1983.

\bibitem[Sky19]{DL4JREPO}
{GitHub - deeplearning4j/dl4j-examples: Deeplearning4j Examples (DL4J, DL4J
  Spark, DataVec)}.
\newblock \url{https://github.com/deeplearning4j/dl4j-examples}, 2019.

\bibitem[SSS{\etalchar{+}}17]{GO}
David Silver, Julian Schrittwieser, Karen Simonyan, Ioannis Antonoglou, Aja
  Huang, Arthur Guez, Thomas Hubert, Lucas Baker, Matthew Lai, Adrian Bolton,
  Yutian Chen, Timothy Lillicrap, Fan Hui, Laurent Sifre, George van~den
  Driessche, Thore Graepel, and Demis Hassabis.
\newblock Mastering the game of go without human knowledge.
\newblock {\em Nature}, 550:354--, 2017.

\bibitem[SSSG83]{Child3}
Prentice Starkey, E~S~Spelke, and Rochel Gelman.
\newblock Detection of intermodal numerical correspondences by human infants.
\newblock {\em Science (New York, N.Y.)}, 222:179--81, 1983.

\bibitem[SSSG90]{Child4}
Prentice Starkey, Elizabeth S.~Spelke, and Rochel Gelman.
\newblock Numerical abstraction by human infants.
\newblock {\em Cognition}, 36:97--127, 1990.

\bibitem[Sta92]{Child2}
Prentice Starkey.
\newblock The early development of numerical reasoning.
\newblock {\em Cognition}, 43:93--126, 1992.

\bibitem[swi19]{SWIFTREPO}
{GitHub - tensorflow/swift-models: Models and examples built with Swift for
  TensorFlow}.
\newblock \url{https://github.com/tensorflow/swift-models}, 2019.

\bibitem[Sza00]{Replicators}
E{\"o}rs Szathm{\'a}ry.
\newblock Evolution of replicators.
\newblock {\em Philosophical transactions of the Royal Society of London.
  Series B, Biological sciences}, 355:1669--1676, 2000.

\bibitem[SZF{\etalchar{+}}17]{DND}
Andr{\'a}s Szil{\'a}gyi, Istv{\'a}n Zachar, Anna Fedor, Harold de~Vladar, and
  E{\"o}rs Szathm{\'a}ry.
\newblock Breeding novel solutions in the brain: A model of darwinian
  neurodynamics.
\newblock {\em F1000Research}, 5(2416), 2017.

\bibitem[Tan19]{TFREPO}
Yuan Tang.
\newblock Github - tensorflow/tensorflow: An open source machine learning
  framework for everyone.
\newblock \url{https://github.com/tensorflow/tensorflow}, 2019.

\bibitem[Tri92]{Child5}
Lana Trick.
\newblock A theory of enumeration that grows out of a general theory of vision:
  Subitizing, counting, and finsts.
\newblock {\em The Nature and Origins of Mathematical Skills}, 91:257--299,
  1992.

\bibitem[Tur36]{Turing1936}
Alan~M. Turing.
\newblock On computable numbers, with an application to the
  {E}ntscheidungsproblem.
\newblock {\em Proceedings of the London Mathematical Society}, 2(42):230--265,
  1936.

\bibitem[VBC{\etalchar{+}}19]{AlphaStarB}
Oriol Vinyals, Igor Babuschkin, Junyoung Chung, Michael Mathieu, Max Jaderberg,
  Wojciech~M. Czarnecki, Andrew Dudzik, Aja Huang, Petko Georgiev, Richard
  Powell, Timo Ewalds, Dan Horgan, Manuel Kroiss, Ivo Danihelka, John Agapiou,
  Junhyuk Oh, Valentin Dalibard, David Choi, Laurent Sifre, Yury Sulsky, Sasha
  Vezhnevets, James Molloy, Trevor Cai, David Budden, Tom Paine, Caglar
  Gulcehre, Ziyu Wang, Tobias Pfaff, Toby Pohlen, Yuhuai Wu, Dani Yogatama,
  Julia Cohen, Katrina McKinney, Oliver Smith, Tom Schaul, Timothy Lillicrap,
  Chris Apps, Koray Kavukcuoglu, Demis Hassabis, and David Silver.
\newblock {AlphaStar: Mastering the Real-Time Strategy Game StarCraft II}.
\newblock
  \url{https://deepmind.com/blog/alphastar-mastering-real-time-strategy-game-starcraft-ii/},
  2019.

\bibitem[VEB{\etalchar{+}}17]{AlphaStar0}
Oriol Vinyals, Timo Ewalds, Sergey Bartunov, Petko Georgiev, Alexander~Sasha
  Vezhnevets, Michelle Yeo, Alireza Makhzani, Heinrich K{\"{u}}ttler, John
  Agapiou, Julian Schrittwieser, John Quan, Stephen Gaffney, Stig Petersen,
  Karen Simonyan, Tom Schaul, Hado van Hasselt, David Silver, Timothy~P.
  Lillicrap, Kevin Calderone, Paul Keet, Anthony Brunasso, David Lawrence,
  Anders Ekermo, Jacob Repp, and Rodney Tsing.
\newblock Starcraft {II:} {A} new challenge for reinforcement learning.
\newblock {\em CoRR}, abs/1708.04782, 2017.

\bibitem[VF04]{BrainNumbers2}
Tom Verguts and Wim Fias.
\newblock Representation of number in animals and humans: A neural model.
\newblock {\em J. Cognitive Neuroscience}, 16(9):1493--1504, 2004.

\bibitem[VLS90]{Child6}
Erik Van~Loosbroek and Ad~Smitsman.
\newblock Visual perception of numerosity in infancy.
\newblock {\em Developmental Psychology}, 26:916--922, 1990.

\bibitem[vN63]{NeumannCW}
John von Neumann.
\newblock The general and logical theory of automata.
\newblock In A.~H. Taub, editor, {\em {John von Neumann: Collected Works.
  Volume V: Design of Computers, Theory of Automata and Numerical Analysis}}.
  Pergamon Press, 1963.

\bibitem[vOGV82]{GaussOFS}
Michiel van Oeffelen and Peter G.~Vos.
\newblock A probabilistic model for the discrimination of visual number.
\newblock {\em Perception \& psychophysics}, 32:163--70, 1982.

\bibitem[WZS18]{DCNNNUM}
Xiaolin Wu, Xi~Zhang, and Xiao Shu.
\newblock Cognitive deficit of deep learning in numerosity.
\newblock {\em CoRR}, abs/1802.05160, 2018.

\bibitem[ZF13]{ConvNet}
Matthew~D. Zeiler and Rob Fergus.
\newblock Visualizing and understanding convolutional networks.
\newblock {\em CoRR}, abs/1311.2901, 2013.

\end{thebibliography}
